# A Readability Analysis of Campaign Speeches from the 2016 US Presidential Campaign

Elliot Schumacher, Maxine Eskenazi

Language Technologies Institute

School of Computer Science
Carnegie Mellon University

5000 Forbes Ave., Pittsburgh, PA 15213

www.lti.cs.cmu.edu

*Introduction*

The goal of this report is to assess the readability of the campaign speeches of five presidential candidates in the 2016 US presidential race and to examine their evolution over time and according to the type of speech. Readability can be defined here as the reading level, from grade 1 to grade 12, of a document. It is determined by looking at the lexical contents and the grammatical structure of the sentences in a document. It is based on the observation that some words (and grammatical structures) appear with greater frequency at one grade level than another. For example, we would expect that we could see the word "win" fairly frequently in third grade documents while the word "successful" would be more frequent in, say, seventh grade documents. We would not see dependent clauses very often at the second grade level whereas they would be quite frequent at the seventh grade level.

For this analysis, we use a readability model, REAP, that was developed for vocabulary at by Collins-Thompson and Callan (2004) and further developed for grammar by Heilman et al (2006, 2007). It is based on a database of sets of texts, one set for each grade level. Most of the texts come from student-written texts that teachers have published on their websites, noting the grade that each represents. The lexical reading difficulty measure is based on the smoothed individual probabilities of words occurring at each reading level. For example, the word, *determine,* was predictive of Grade 11 text, and was more predictive of high school-level text than lower-level text. The grammar reading difficulty measure is based on the one- to three-level depth parse trees of the sentences. This means that the measure is based on typical grammatical constructions in sentences of each grade level.

*Background*

Early readability measures made assumptions about what a difficult text was. The Dale-Chall Readability Formula (Dale and Chall, 1948) defined the readability level as a linear function of the average number of words in a sentence and the percentage of rare words in the document. Flesch-Kincaid (Kincaid et al 1975) was based on the average sentence length and the average number of syllables per word.

More recently, the Lexile Framework (version 1.0, Stenner, 1996) uses word frequency estimates as a measure of lexical difficulty and sentence length as a grammatical feature. Other approaches characterized text in more holistic terms. Coh-Metrix (Graesser et al 2011) measures text cohesiveness, accounting for both the reading difficulty of the text and other lexical and syntactic measures as well as a measure of prior knowledge needed for comprehension and the genre of the text. These factors account for the difficulty of constructing the mental representation of the text.

All of the measures, REAP included, were originally developed to help teachers choose appropriate documents for their students in reading classes. The campaign speeches, while most were written in advance, are destined to be spoken. Written speech is very different from spoken speech. When we speak we usually use less structured language with shorter sentences. So while measures such as Flesch-Kincaid are appropriate for written speech, they are not really reflective of the structure of spoken language. REAP has been trained on written texts, as described above. But it concentrates on how often words and grammatical constructs are used at each grade level and less on the length of the sentence and of each word. So REAP corresponds better to an analysis of spoken language than its predecessor.

*Methodology*

A database was collected containing documents from each of the five current presidential candidates: Ted Cruz (5), Hillary Clinton (7), Marco Rubio (6), Bernie Sanders (6), Donald Trump (8) (see References and Appendix). The documents are transcriptions of their campaign speeches. They range from the declaration of candidacy speech to campaign trail speeches to victory speeches to defeat speeches. The numbers show it was sometimes difficult to find transcriptions rather than videos. In the future an Automatic Speech Recognition system (ASR) could be used to obtain text from the videos. Given that this process would produce some error, it was not used for the present study. For comparison we also analyzed the readability of Lincoln's Gettysburg Address (Bliss version) and a speech from Barack Obama, George W. Bush, Bill Clinton and Ronald Reagan (the latter two at the same venue in different years).
Two levels of analysis were carried out. First we looked at level just based on the vocabulary content. The second analysis looked at syntax structure.

*Results*

Figure 1 shows that speeches by past presidents while on campaign and the Gettysburg Address were at least at the eighth grade level. The candidates' speeches mostly went from seventh grade level for Donald Trump to tenth grade level for Bernie Sanders.

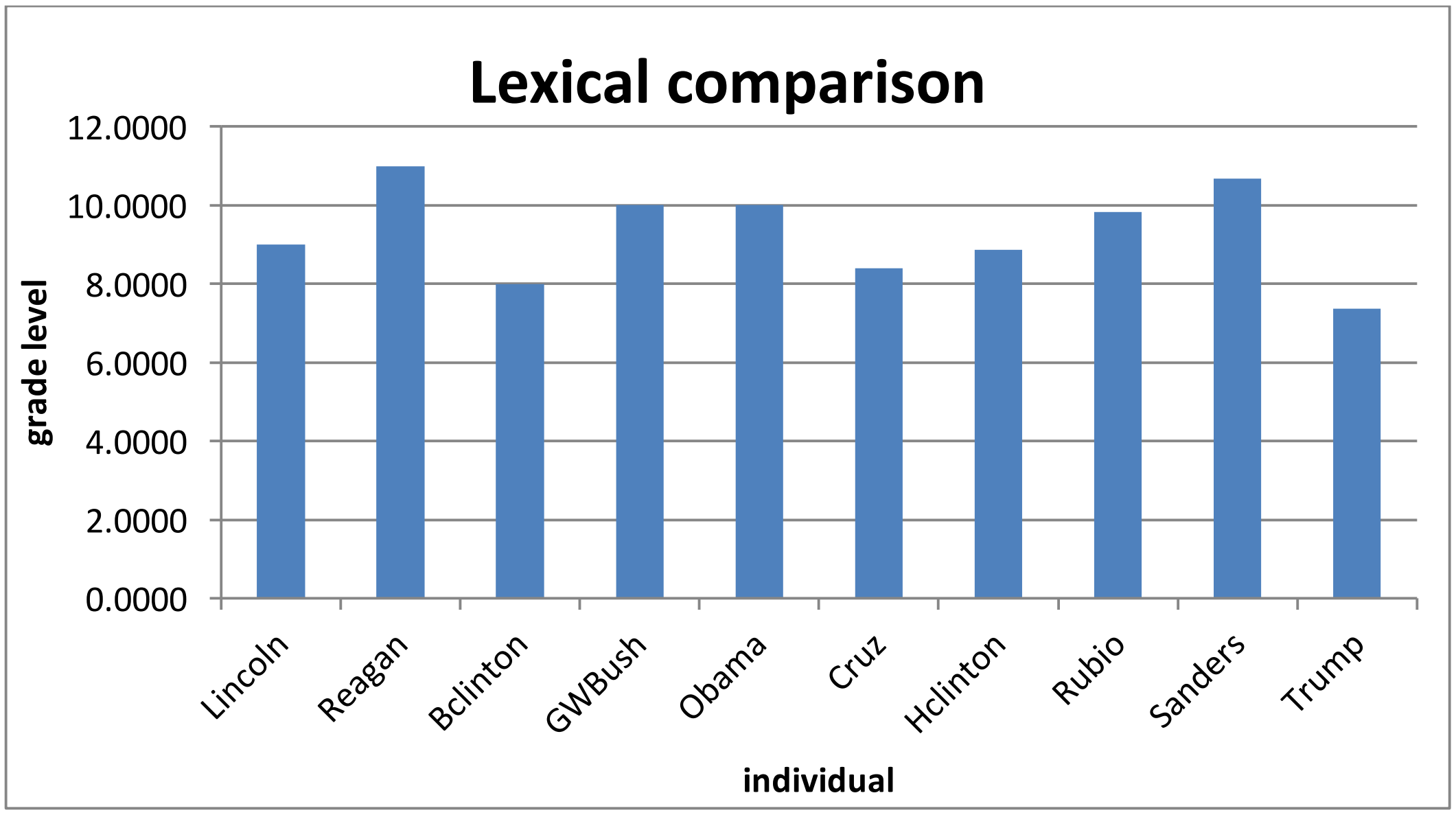

Figure 1. REAP lexical measure

We can compare this to the analysis carried out by the Boston Globe (Boston Globe) using the Flesch-Kincaid measure on the candidates' 2015 speeches as shown in Figure 2. They performed their analysis only on each candidate's campaign announcements.

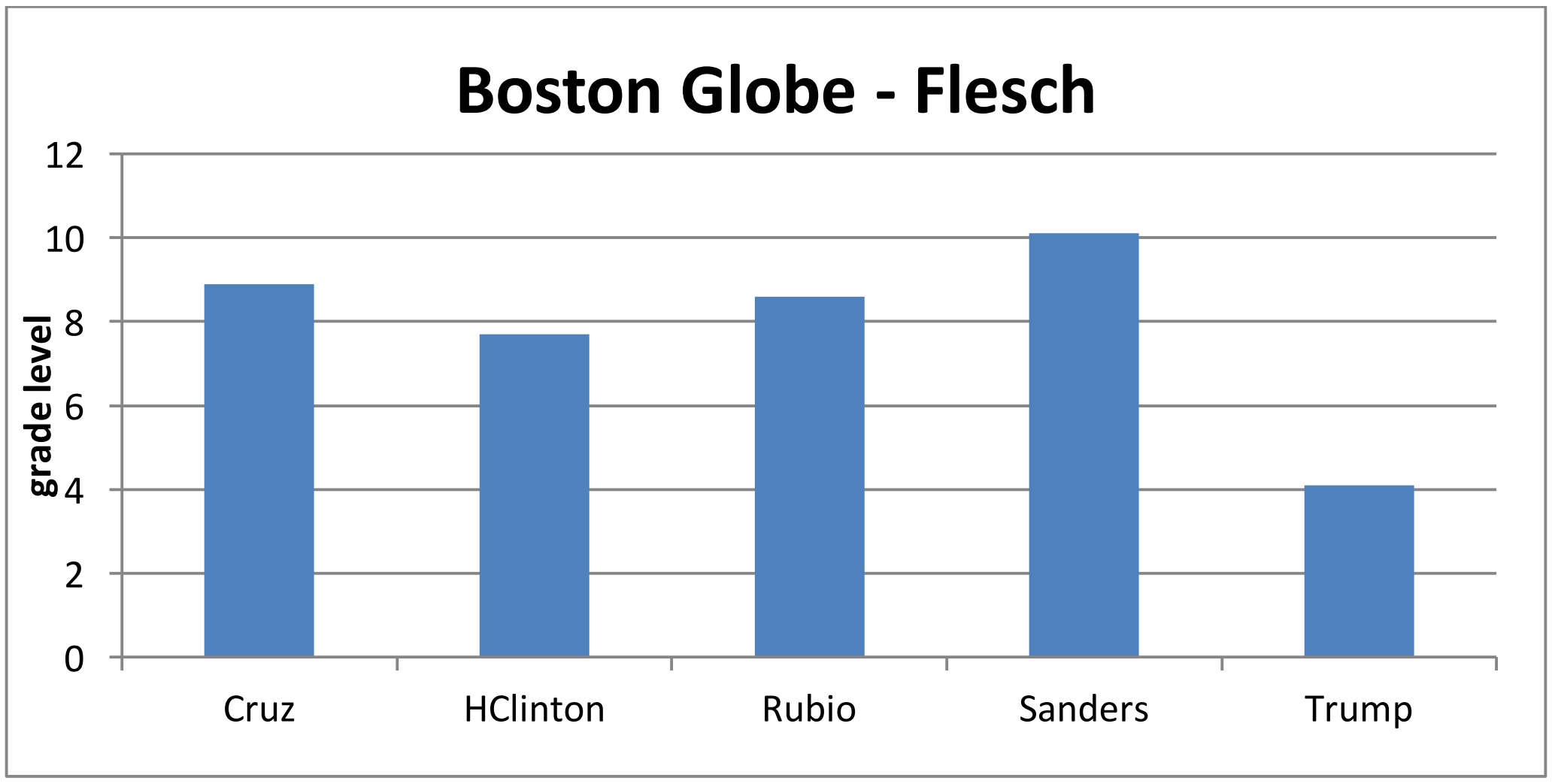

Figure 2. Boston Globe Flesch-Kincaid measures for 2015 campaign speeches

It would appear that an analysis more geared toward spoken language gives both Mr. Trump and Mrs. Clinton higher scores for their choice of words.

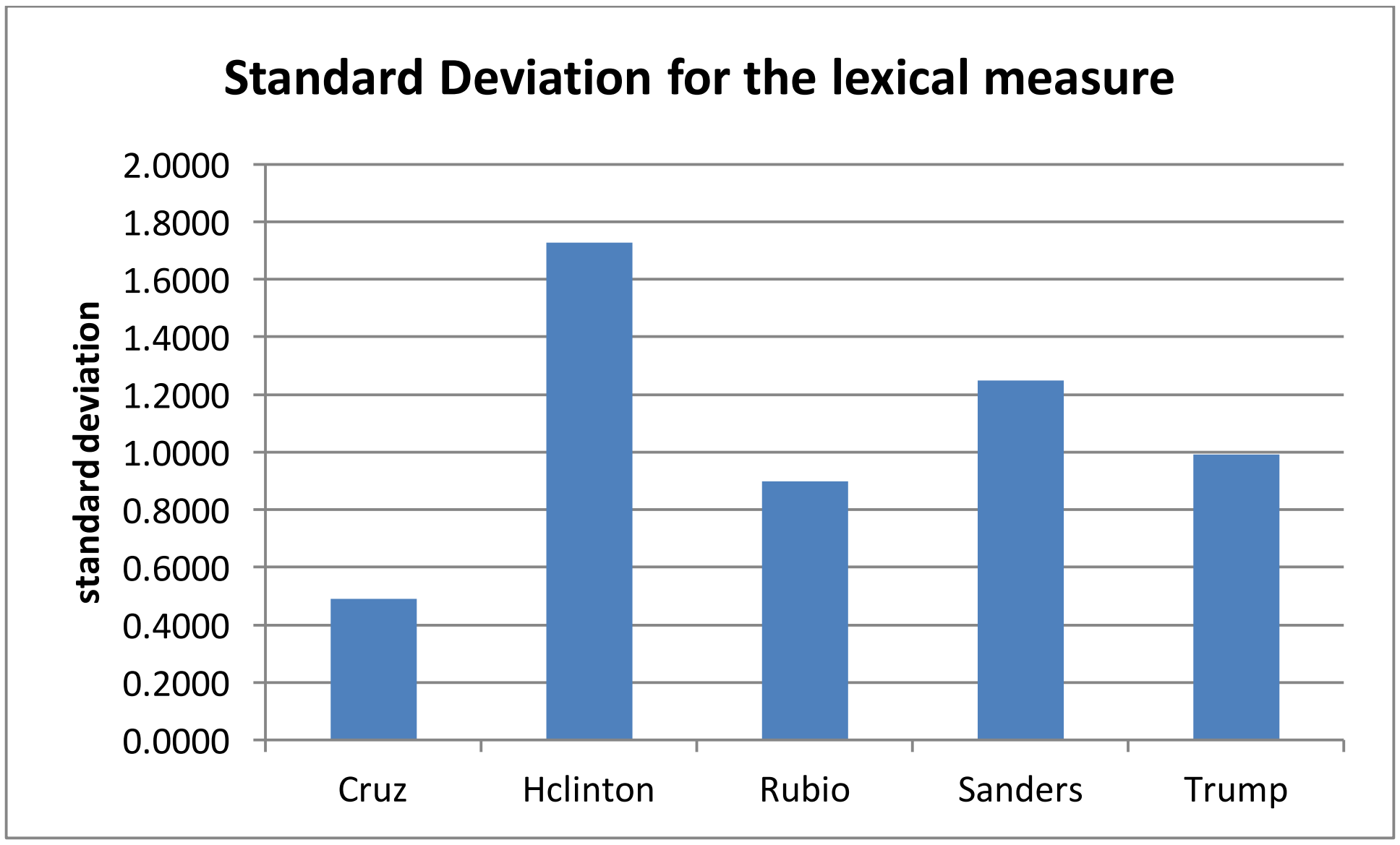

Figure 3. REAP lexical measure standard deviation per candidate

Figure 3 shows the standard deviation of the scores in Figure 1. This reveals the degree to which the candidate changes their choice of words from one speech to another. This could reflect an effort to take into account the different audiences or circumstances (winning or concession speech in a state, for example). We can see that Hilary Clinton has the highest standard deviation and so the biggest change of choice of words from one speech to another, while Ted Cruz varies the least in his choices.

We also compared the grammar levels for all of the candidates and past presidents as shown in Figure 4.

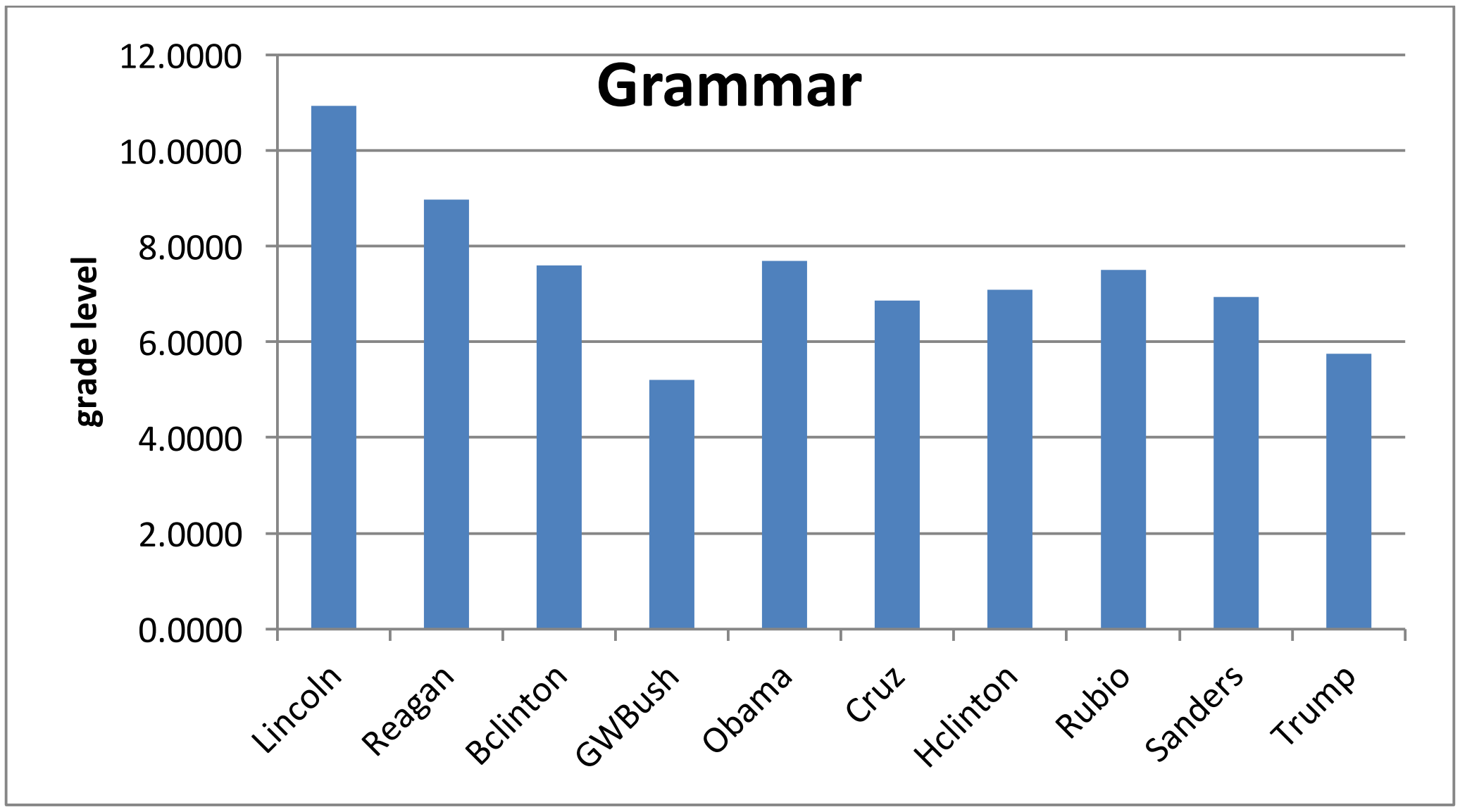

Figure 4. REAP grammar measure

We see that George W. Bush had the lowest level and Abraham Lincoln the highest. Amongst the candidates, levels are between sixth and seventh grades except for Donald Trump (grade 5.7).

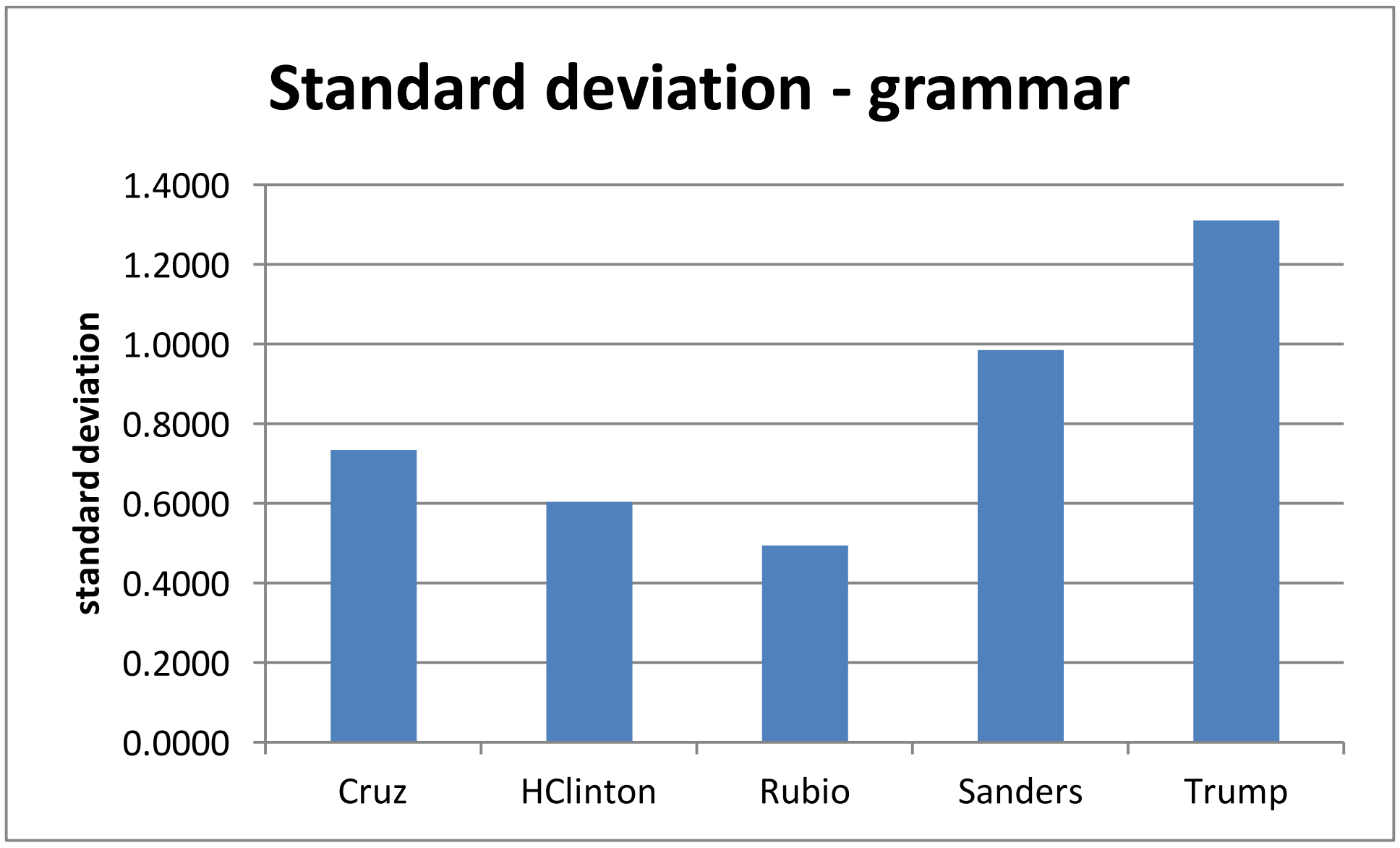

Figure 5. Grammar standard deviation

Looking at the standard deviation of the candidates on the grammar level, Donald Trump stands out as having the greatest change in the structure of his speeches while Marco Rubio has the lowest level of variation.

Candidates give speeches to differing types of audiences over time, ranging from small gatherings with a specific issue in mind to larger general ones. The one speech made by every one of the candidates was the *announcement of candidacy*. Figure 6 shows the lexical level of these speeches and Figure 7 shows the grammar level. We note that lexical levels are comparable for most candidates with Donald Trump and Hilary Clinton having the lowest levels, at grade 8. For grammar, we see that the level for Donald Trump is significantly lower, at grade 5.

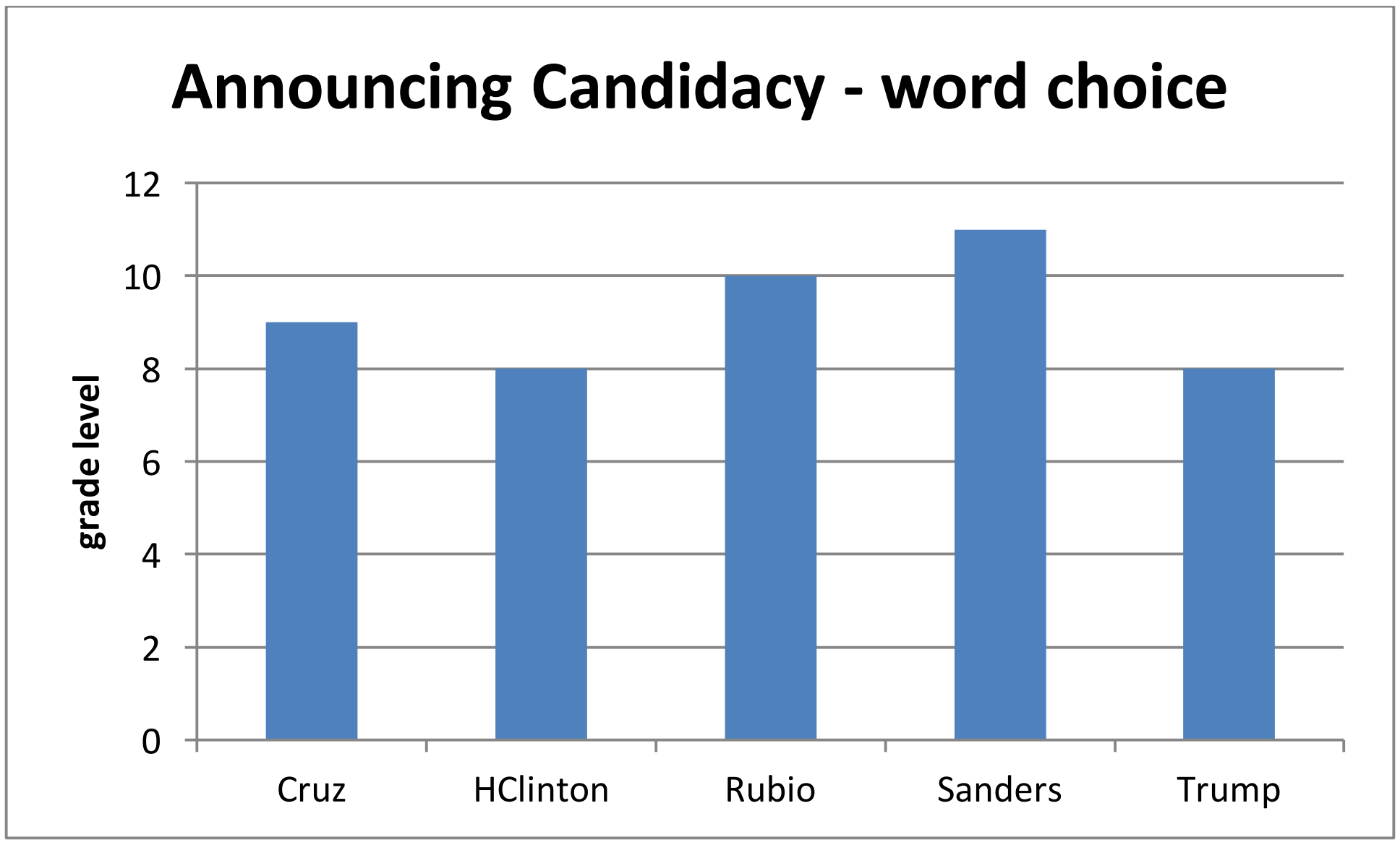

Figure 6. Lexical level of candidacy announcement speeches

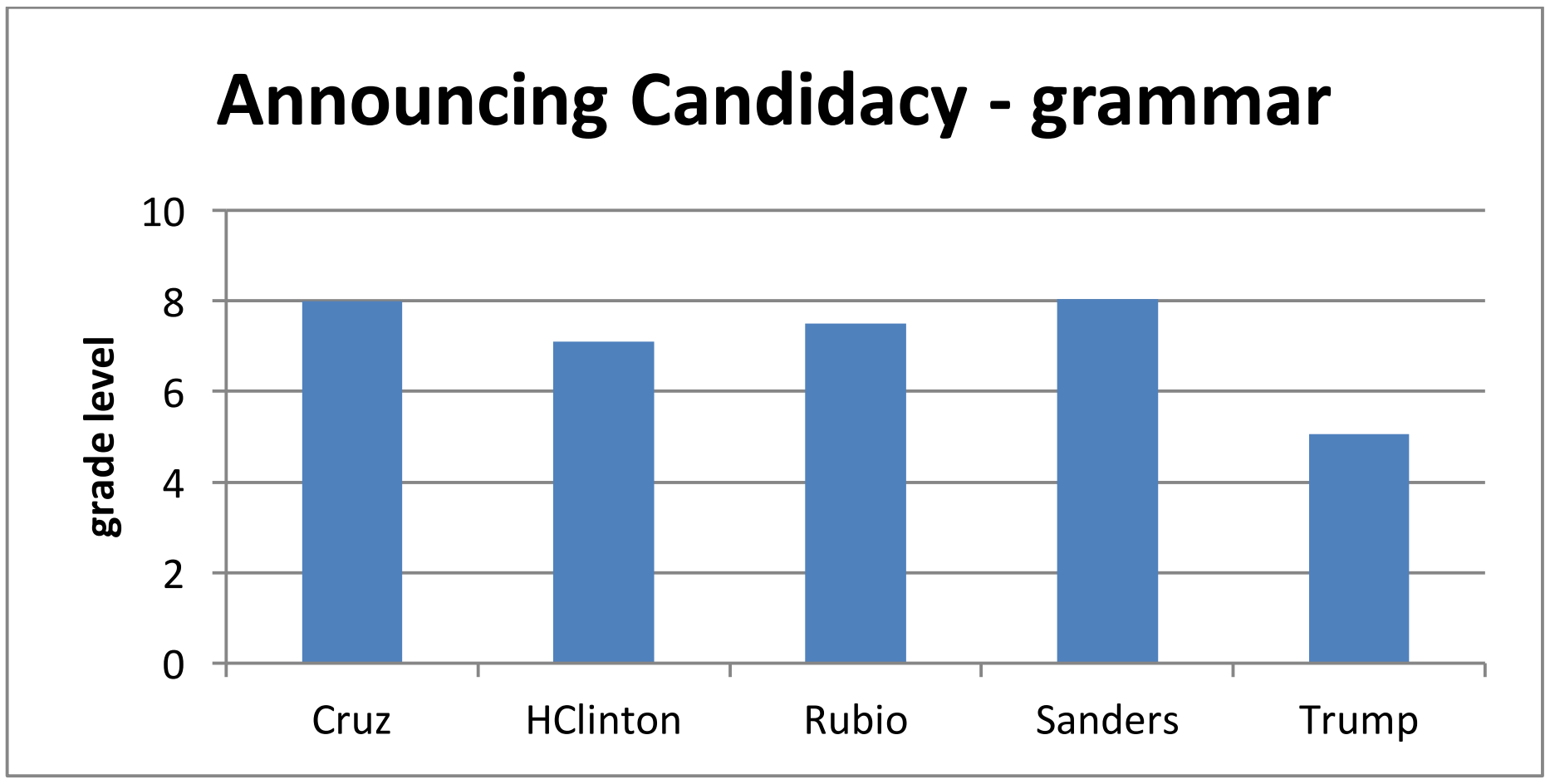

Figure 7. Grammar level of candidacy announcement speeches

Finally, we looked at whether the levels of the speeches had varied over time. Figures 8, 9, 10, 11 and 12 show the variation of levels for the five candidates. We also show the variation in the level of grammar in Figures 13, 14, 15, 16 and 17. It should be noted that although video is generally available for all of the candidates' speeches, transcripts are not as readily available. With the exception of the candidacy speech, we did not find one same venue for the all of the candidates. We note here that we voluntarily did not look at the transcriptions of the debates (if available), which would produce similar settings for all of the candidates of the same party. Nor did we find transcriptions for all of the candidates on one same date.

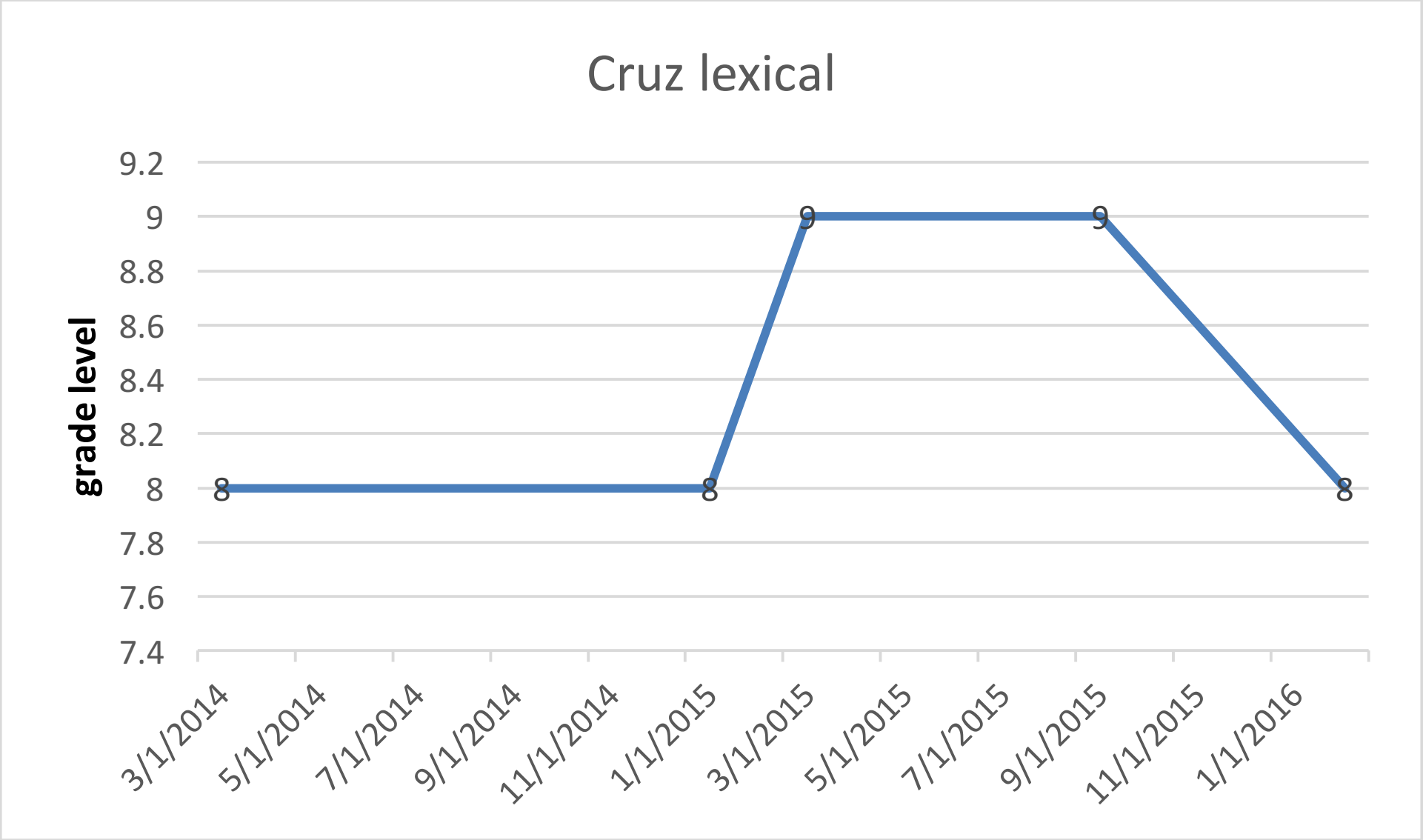

Figure 8. Evolution of lexical level over time – Cruz

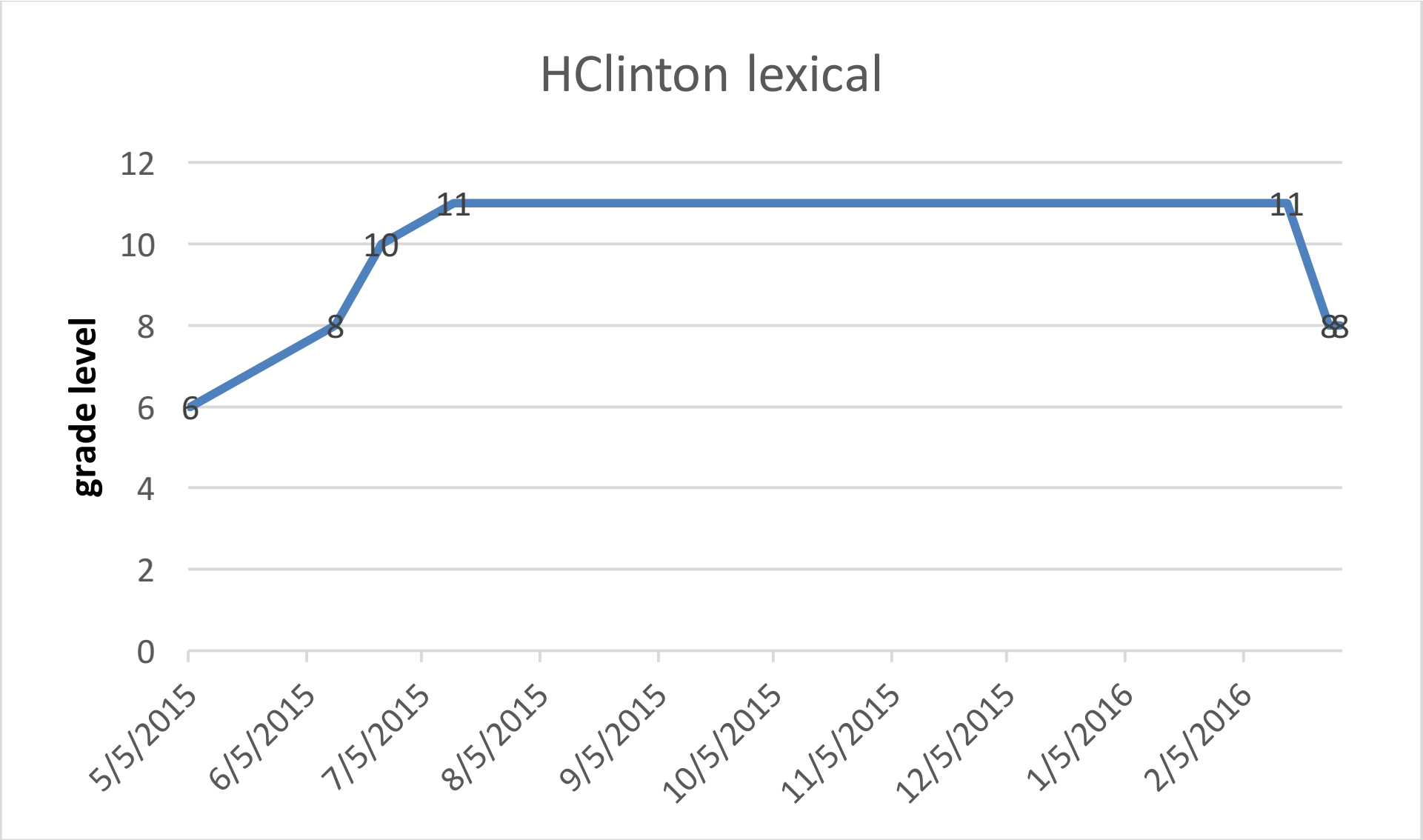

Figure 9. Evolution of lexical level over time – H Clinton

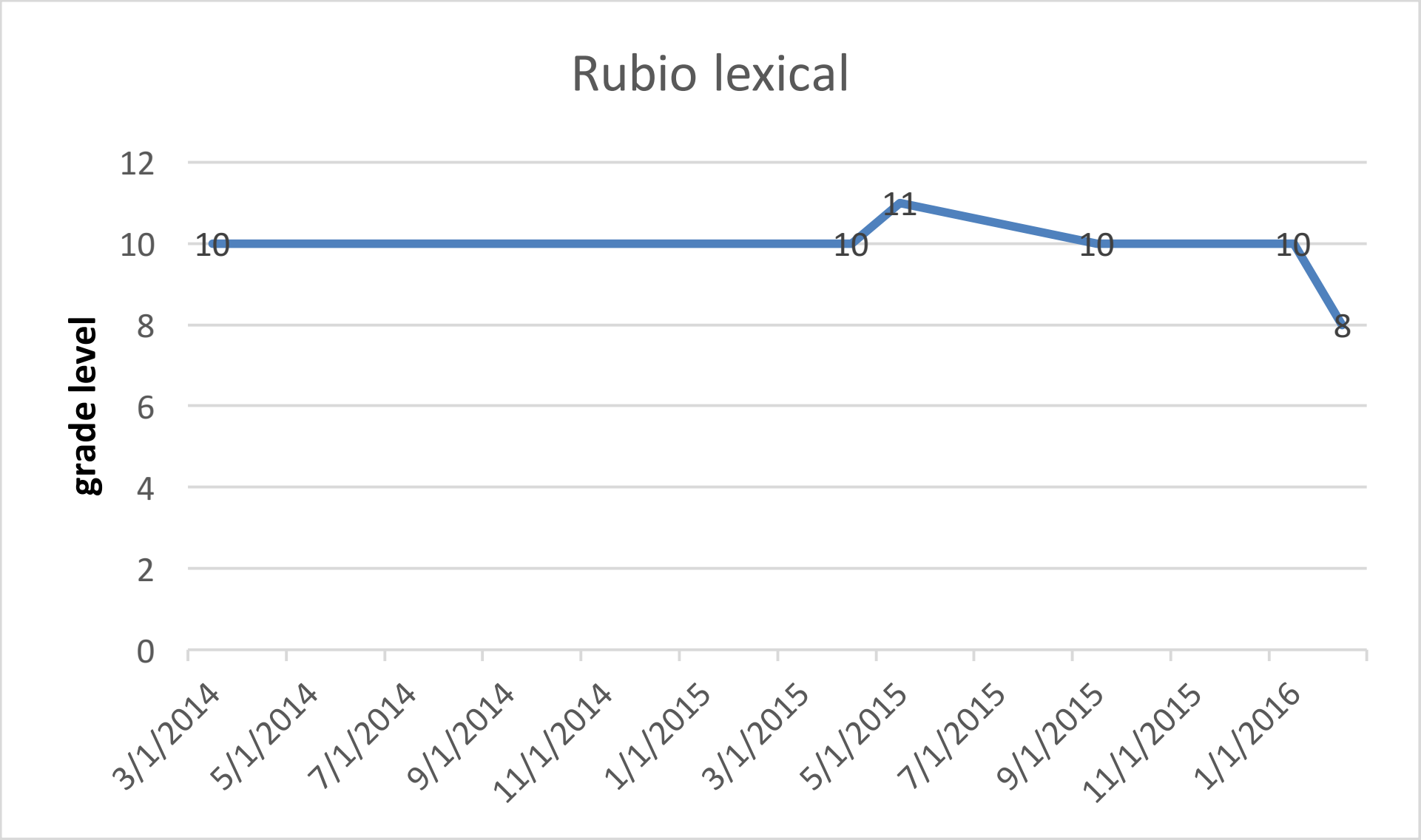

Figure 10. Evolution of lexical level over time – Rubio

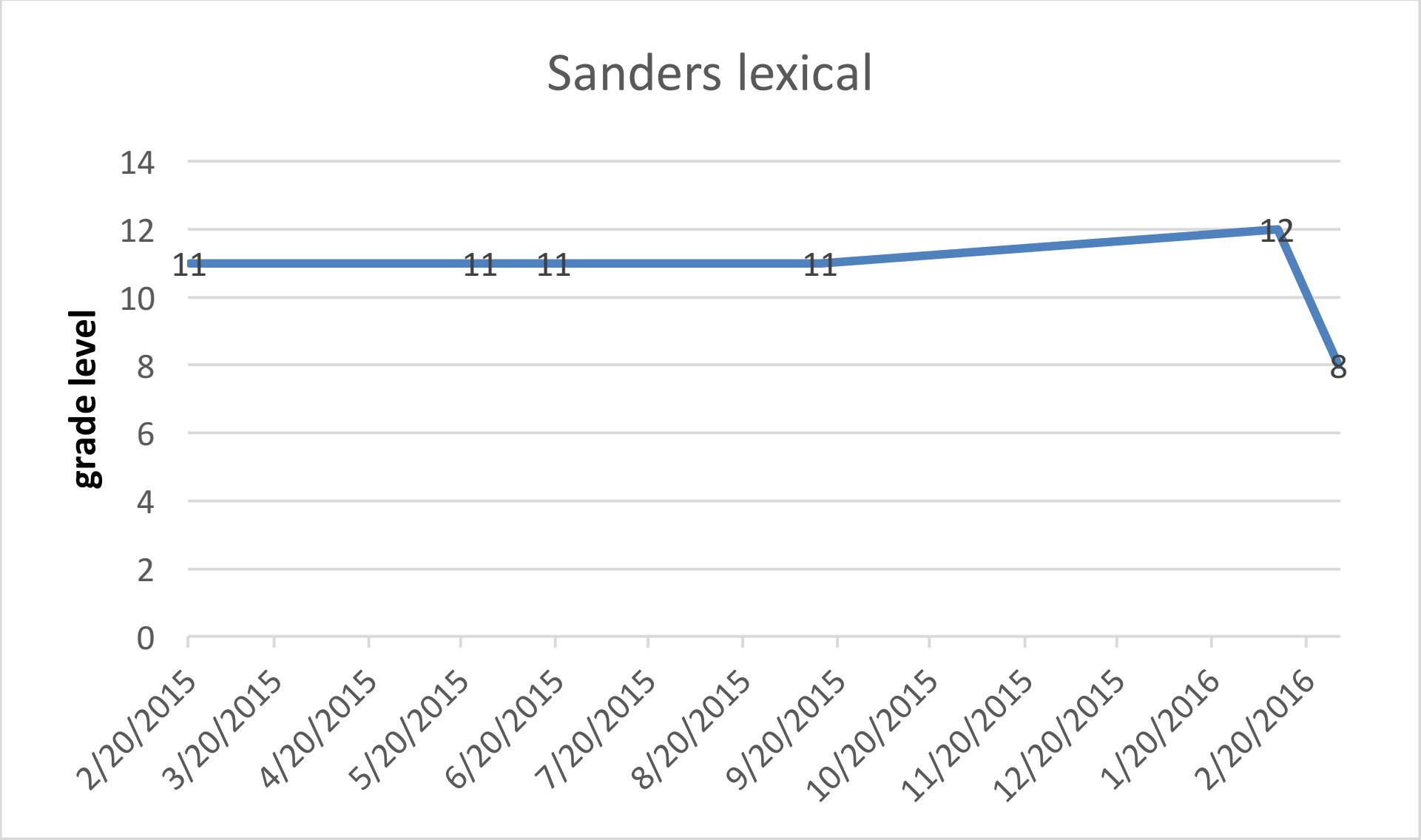

Figure 11. Evolution of lexical level over time – Sanders

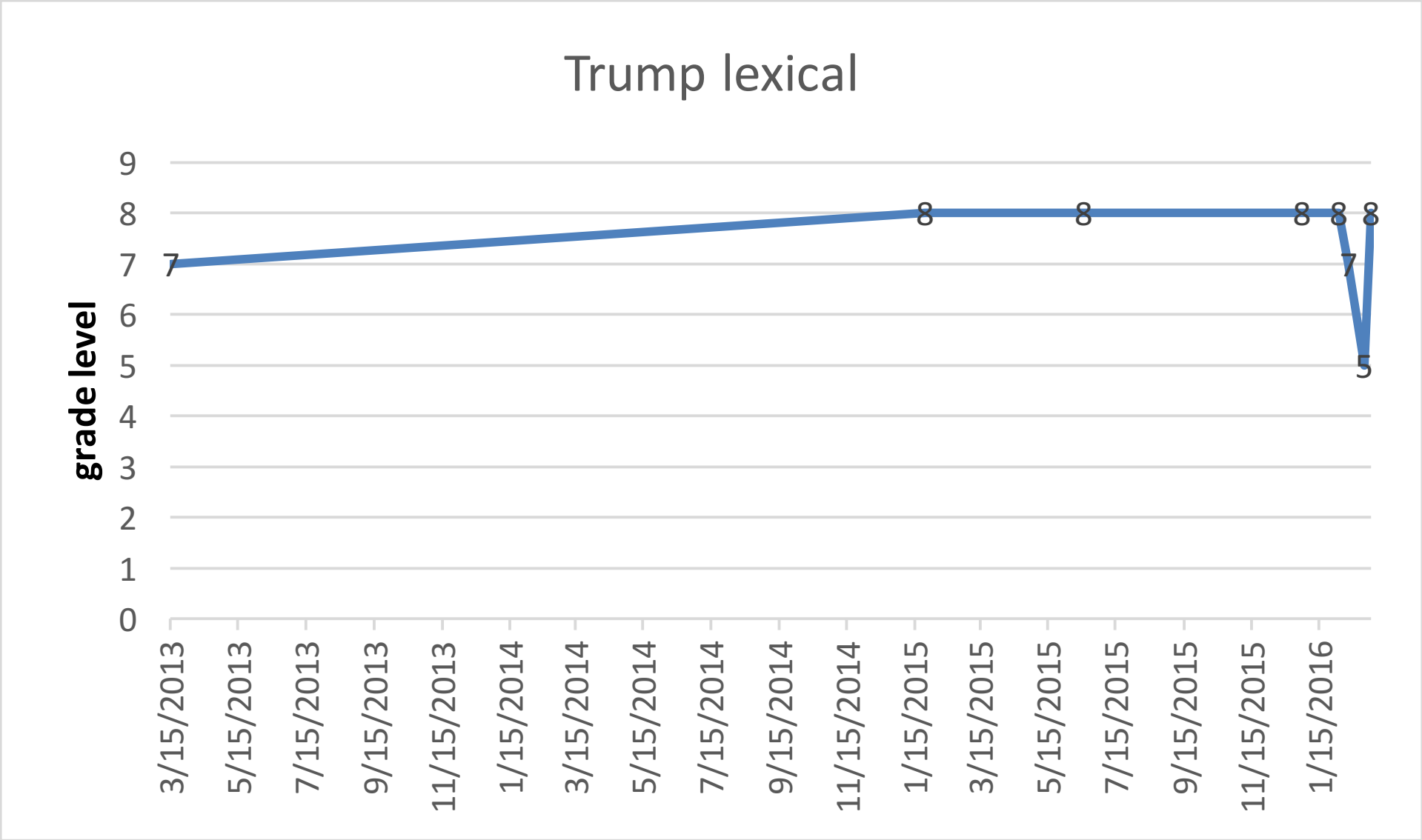

Figure 12. Evolution of lexical level over time – Trump

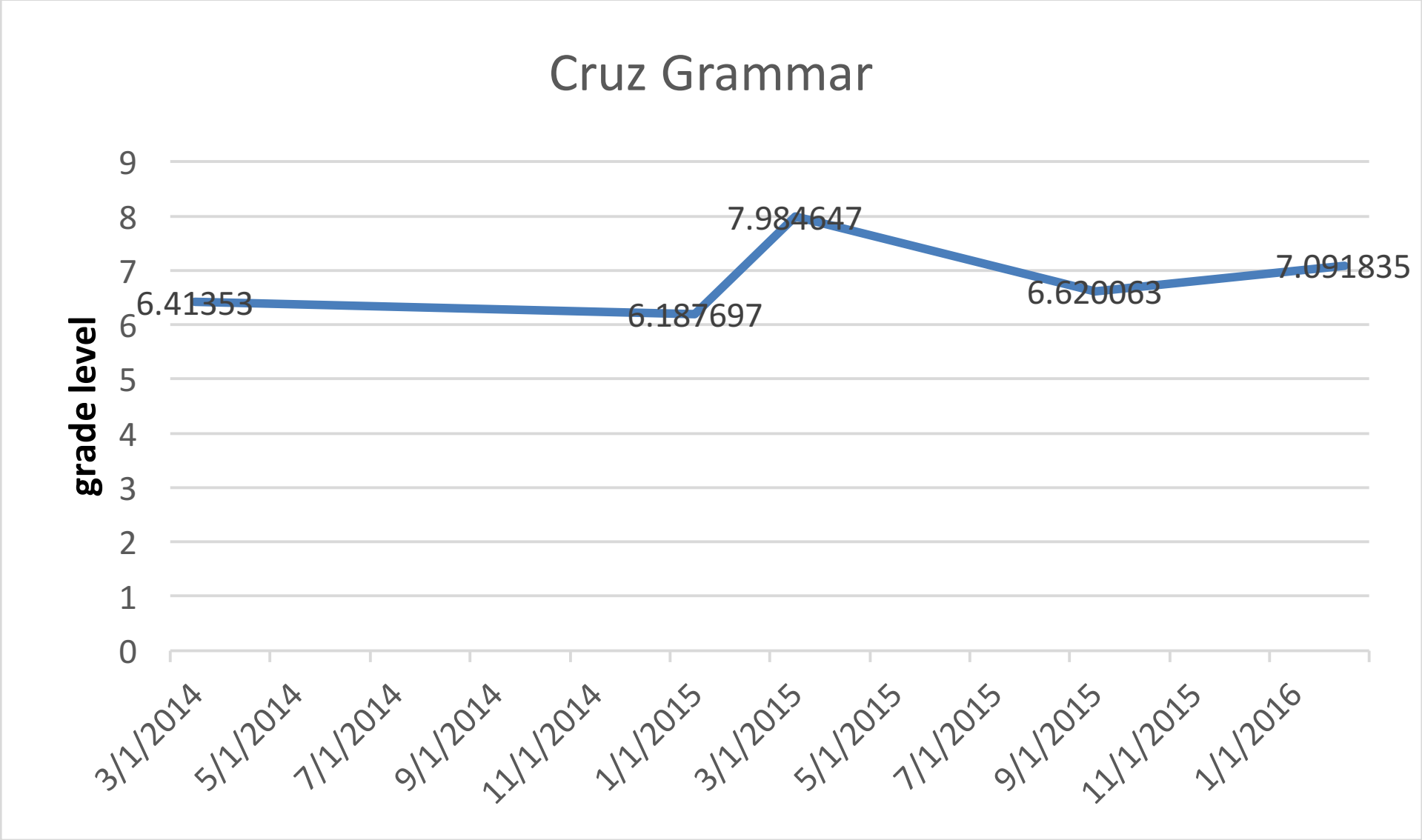

Figure 13. Evolution of grammar level over time – Cruz

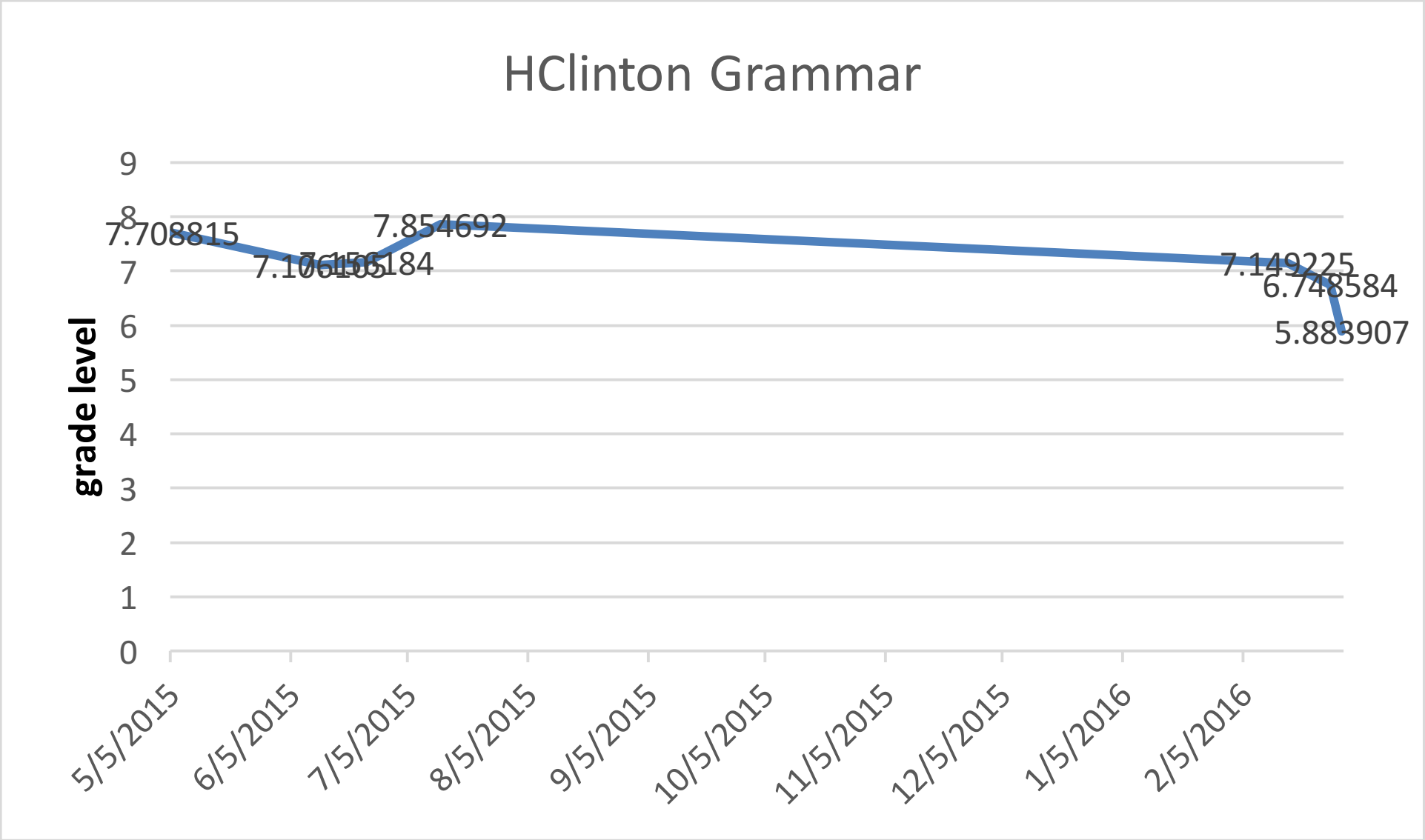

Figure 14. Evolution of grammar level over time – H Clinton

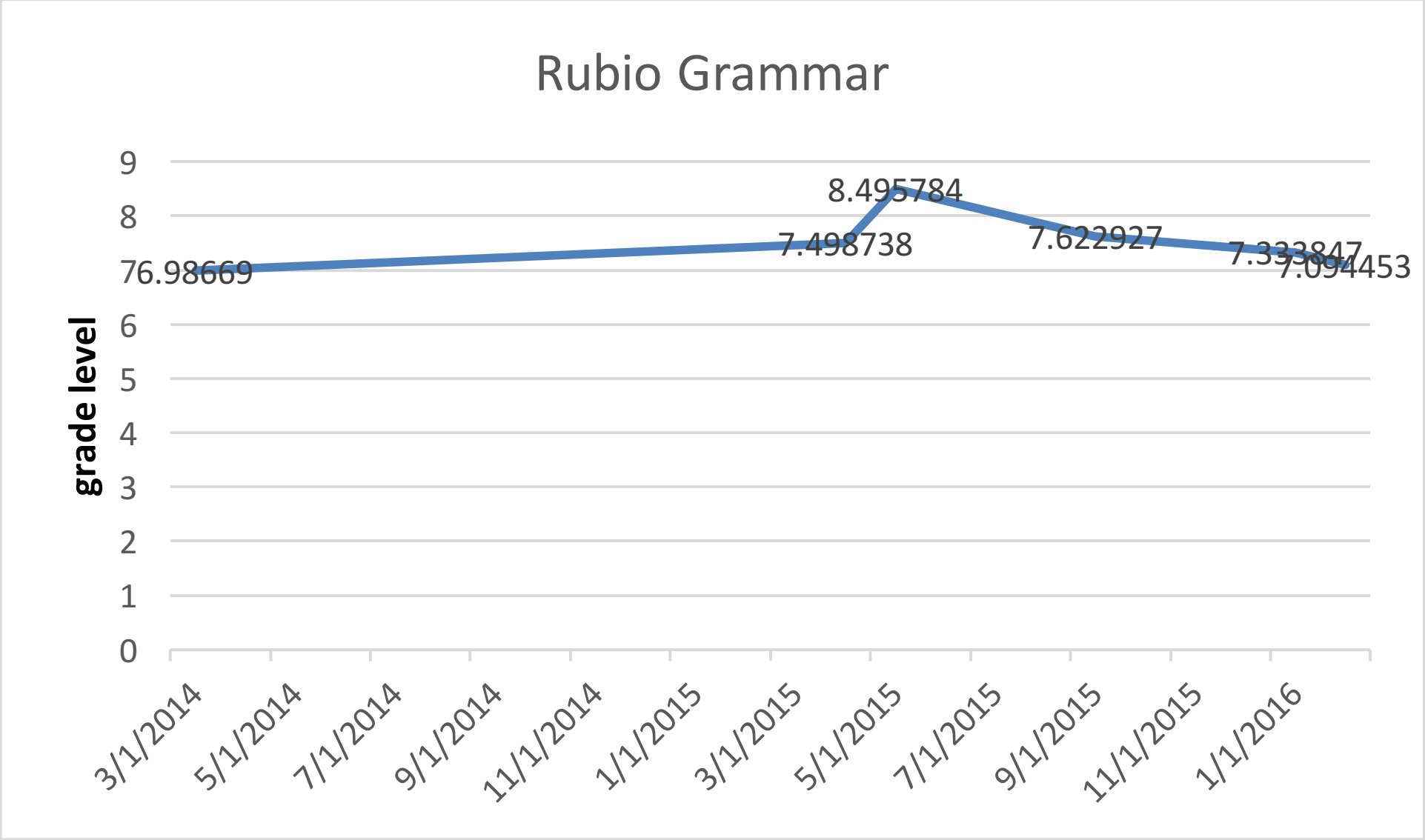

Figure 15. Evolution of grammar level over time – Rubio

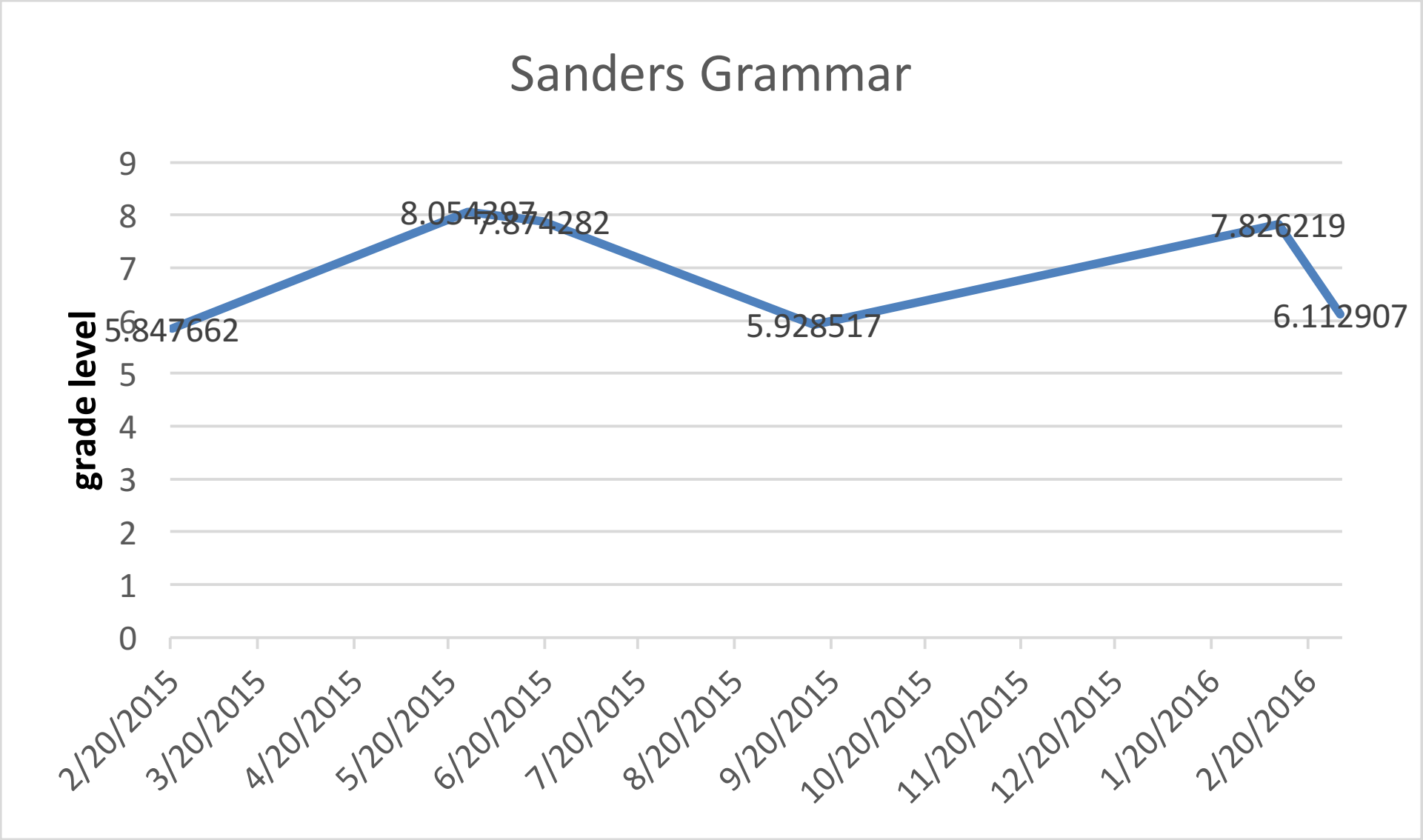

Figure 16. Evolution of grammar level over time – Sanders

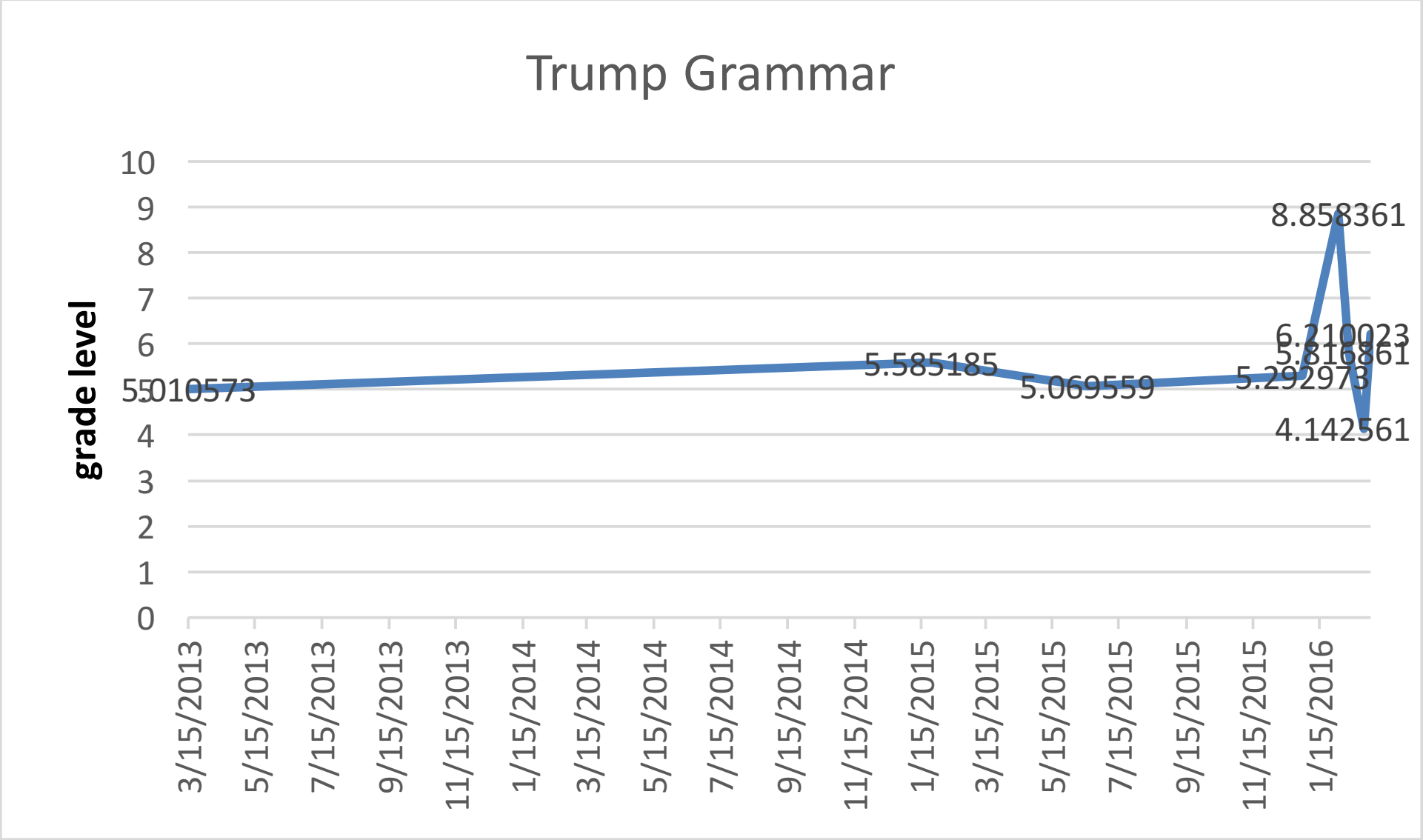

Figure 17. Evolution of grammar level over time – Trump

The results do not show a marked trend over time for any of the candidates, except for the upward trend for Hilary Clinton after her first two speeches. There are a few peaks and valleys worthy of note. First, some measures seem to be lower for the candidates' latest speech. There is also an interesting peak for grammar for Donald Trump in his Iowa concession speech and a considerably lower level of both lexicon and grammar for Trump for his Nevada victory speech (while the same is not seen for his Super Tuesday victory speech).

*Conclusions*

This technical report has assessed the lexical and grammatical levels of the 2016 presidential candidates' speeches. This analysis shows the changes that candidates make in the level of their speech according to the type of speech. It also reflects each candidate's combination of personal delivery style and their analysis of the level of the audience they want to address.

Speeches

Appendix – List of Candidates' speeches

| Candidate | Date | Occasion | Grammar | Lexical |
|---|---|---|---|---|
| Cruz | 1/24/2015 | Iowa Freedom Summit | 6.187697 | 8 |
| Cruz | 3/23/2015 | Campaign Announcement - Liberty University | 7.984647 | 9 |
| Cruz | 2/1/2016 | Iowa Caucus Election Night | 7.091835 | 8 |
| Cruz | 9/25/2015 | 2015 Values Voter Summit | 6.620063 | 9 |
| Cruz | 3/7/2014 | CPAC 2014 | 6.41353 | 8 |
| Hclinton | 5/5/2015 | Town Hall Immigration in Nevada | 7.708815 | 6 |
| Hclinton | 6/12/2015 | Campaign Announcement | 7.106105 | 8 |
| Hclinton | 6/24/2015 | Speech in Missouri Church | 7.156184 | 10 |
| Hclinton | 7/13/2015 | Economic Speech at New School | 7.854692 | 11 |
| Hclinton | 2/16/2016 | Schomburg Center for Research in Black Culture in Harlem, New York | 7.149225 | 11 |
| Hclinton | 2/27/2016 | South Carolina Victory Speech | 6.748584 | 8 |
| Hclinton | 3/1/2016 | Super Tuesday Victory Speech | 5.883907 | 8 |
| Rubio | 3/6/2014 | CPAC 2014 | 6.98669 | 10 |
| Rubio | 4/13/2015 | Campaign Announcement | 7.498738 | 10 |
| Rubio | 5/21/2015 | Council on Foreign Relations | 8.495784 | 11 |
| Rubio | 9/25/2015 | Value Voters Summit 2015 | 7.622927 | 10 |
| Rubio | 1/4/2016 | Speech in New Hampshire | 7.333847 | 10 |
| Rubio | 2/20/2016 | South Carolina Election Night | 7.094453 | 8 |
| Sanders | 2/20/2015 | Nevada Election Night Speech | 5.847662 | 11 |
| Sanders | 5/26/2015 | Campaign Announcement | 8.054397 | 11 |
| Sanders | 6/19/2015 | NALEO Conference | 7.874282 | 11 |
| Sanders | 9/14/2015 | Liberty University | 5.928517 | 11 |
| Sanders | 2/10/2016 | New Hampshire Election Night | 7.826219 | 12 |
| Sanders | 3/1/2016 | Super Tuesday Victory Speech | 6.112907 | 8 |
| Trump | 3/15/2013 | CPAC 2013 | 5.010573 | 7 |
| Trump | 1/24/2015 | Iowa Freedom Summit | 5.585185 | 8 |
| Trump | 6/16/2015 | Campaign Announcement | 5.069559 | 8 |
| Trump | 12/30/2015 | S.C. Campaign Speech | 5.292973 | 8 |
| Trump | 2/1/2016 | Iowa Caucus Election Night | 8.858361 | 8 |
| Trump | 2/10/2016 | NH Victory Speech | 5.816861 | 7 |
| Trump | 2/24/2016 | Nevada Victory Speech | 4.142561 | 5 |
| Trump | 3/1/2016 | Super Tuesday Victory Speech | 6.210023 | 8 |